\documentclass{article}

\usepackage[dblblindworkshop, final]{neurips_2026}
\workshoptitle{2nd Workshop on Advances in Representation Learning for Earth Observation}

\usepackage[utf8]{inputenc}
\usepackage[T1]{fontenc}
\usepackage{hyperref}
\usepackage{url}
\usepackage{booktabs}
\usepackage{amsmath,amsfonts}
\usepackage{graphicx}
\usepackage{microtype}
\usepackage{xcolor}
\hypersetup{colorlinks=true,linkcolor=blue,citecolor=blue,urlcolor=blue}

\title{Shortcut Learning in Geospatial Foundation Model Embeddings for Cross-Disaster Damage Assessment}

\author{%
  Ali Shibli, Andrea Nascetti, Yifang Ban \\
  Division of Geoinformatics\\
  KTH Royal Institute of Technology\\
  Stockholm, Sweden \\
  \texttt{\{shibli,nascetti,yifang\}@kth.se} \\
}

\begin{document}
\maketitle

\begin{abstract}
Geospatial foundation models can support disaster damage assessment, but high accuracy on familiar disasters may hide shortcut learning: prediction from event or scene characteristics rather than evidence that transfers to a new disaster. We study embeddings from TerraMind, Prithvi-EO-2.0, DOFA, and DINOv3 on the multi-disaster datasets xView2 and BRIGHT. Holding out an entire disaster, instead of randomly splitting image tiles, reduces mean balanced accuracy from 0.807 to 0.638 on xView2 and from 0.818 to 0.568 on BRIGHT. Four sets of tests examine what information supports prediction: event and acquisition context, before/after image inputs, building count, and correct image pairing. Pre-event images remain predictive, and a classifier given only the annotated number of buildings reaches balanced accuracy of 0.629 and 0.653. Combining before/after features performs best on average on xView2, but does not improve over pre-only features on BRIGHT. Shuffling image pairs within the same disaster and damage class causes modest losses. Motivated by these findings, we test whether feature transformations learned only from training disasters improve generalization to unseen disasters. These transformations improve performance, with supervised contrastive learning performing best on xView2 and PCA on BRIGHT. Our results highlight the importance of evaluation on unseen domains for assessing model robustness and show that simple feature-level transformations can improve generalization. Code and experiments can be found at \url{https://github.com/alishibli97/gfm-shortcut-learning}.
\end{abstract}

\section{Introduction}

Geospatial foundation models (GFMs) provide reusable Earth-observation embeddings when downstream labels are scarce. For disaster response, however, a useful representation must transfer to an event that was not present during training. A random image-level split does not test this: tiles from the same hurricane, earthquake, or wildfire can occur on both sides of the split, sharing geography, sensor properties, building density, and the proportion of tiles labeled as damaged.

This setting encourages \emph{shortcut learning}: using an easy correlation that predicts benchmark labels but is not reliable evidence of damage under the intended deployment shift \citep{Geirhos2020Shortcut}. A model may recognize a known event from geography, hazard type, sensor characteristics, land cover, or urban form, then associate it with that event's fraction of damaged tiles. It may also use either image separately without comparing the same location before and after the disaster. Such context can be legitimate; the central question is whether the resulting prediction transfers to an unseen event.

We study embeddings from TerraMind, Prithvi-EO-2.0, DOFA, and DINOv3 on the multi-disaster datasets xView2/xBD \citep{Gupta2019xBD} and BRIGHT \citep{Chen2025BRIGHT}. We compare a random image-level split, denoted IID (independent and identically distributed) because both sets contain the same events, with leave-one-event-out (LOEO), where the test disaster is unseen. Our audit separates four questions: whether prediction uses (i) event prevalence or acquisition metadata, (ii) one temporal view, (iii) annotation-derived building count, or (iv) the correct same-location pre/post match. We then test whether source-only PCA, a supervised MLP, supervised contrastive learning (SupCon) \citep{Khosla2020SupCon}, and cross-event SupCon improve unseen-event prediction. No held-out-event sample is used for training or feature transformation.

Our contributions are:
\begin{itemize}
    \item an event/context audit showing that random splits overestimate transfer and that event prevalence, acquisition metadata, and building count predict damage;
    \item a complete temporal-view and pairing audit showing dataset-dependent use of pre-, post-, and change information, with only modest losses after correspondence is broken; and
    \item source-only controls showing that event--class reweighting has little effect, while compression and supervised representation learning improve prediction on held-out disasters.
\end{itemize}

\section{Related Work}

\paragraph{Robust EO representations}
Prithvi-EO-2.0, TerraMind, and DOFA provide reusable EO representations with different pre-training and sensor designs \citep{Szwarcman2025PrithviEO2,TerraMind2025,Xiong2024DOFA}; DINOv3 provides a large self-supervised vision baseline \citep{Simeoni2025DINOv3}. Robustness benchmarks nevertheless show sensitivity to real distribution shifts \citep{EarthShift2026,REOBench2025}. We focus on a concrete operational shift: training on known disasters and testing on a different disaster.

\paragraph{Shortcut learning and cross-disaster transfer}
Shortcuts are predictive but potentially unstable signals \citep{Geirhos2020Shortcut,Ye2024Spurious}. xBD established paired optical damage assessment, while BRIGHT adds globally distributed optical--SAR pairs \citep{Gupta2019xBD,Chen2025BRIGHT}. Cross-disaster adaptation has also been studied directly \citep{Mouradi2026CrossDisaster}. 
% Our contribution is a targeted audit of which non-transferable signals are available in foundation-model features, followed by simple source-only controls. The contrastive baselines build on supervised and cross-domain contrastive learning \citep{Khosla2020SupCon,Wang2021CDCL}; we do not claim a new loss.
We complement this work with controlled tests of available context and image-pair information. Our transfer baselines use supervised and cross-domain contrastive learning \citep{Khosla2020SupCon,Wang2021CDCL}; our contribution is the evaluation and audit, not a new training loss.
\section{Shortcut Audit and Source-Only Mitigation}

\paragraph{Task and representation}
For backbone $\phi_m$, we independently encode the pre- and post-event images and form
\begin{equation}
z_i=[z_i^{\rm pre};z_i^{\rm post};z_i^{\rm post}-z_i^{\rm pre}],\quad
z_i^t=\phi_m(x_i^t).
\end{equation}
The binary label $y_i$ is one when the tile contains at least one damaged building; $e_i$ is its disaster event. A class-balanced linear probe (linear classifier) predicts $y_i$. We compare a random image-level split (IID), where the same events occur in training and test sets, with LOEO. In LOEO, all samples from one event are held out, and scaling, transformation, and probe fitting use only the remaining source events. Metrics are averaged over events so large disasters do not dominate.

\begin{figure}[t]
    \centering
    \includegraphics[width=\linewidth]{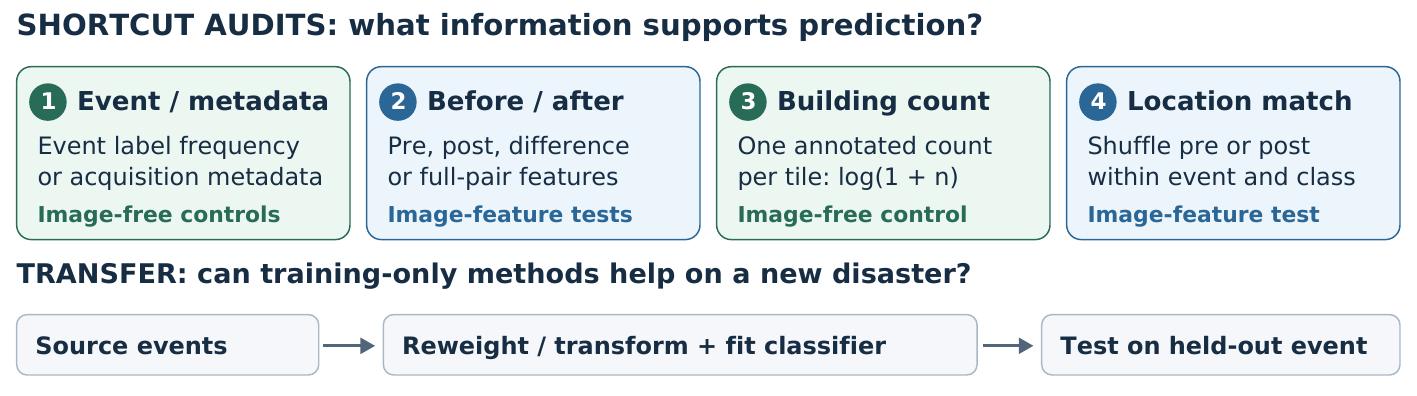}
    \caption{Four questions guide the audit. Image-free controls use metadata or annotations; image tests use fixed-model features. Transfer methods are fitted on source events only.}
    \label{fig:protocol}
\end{figure}

% \paragraph{What shortcuts are available?}
% We use four diagnostic families. \textbf{(1) Event/context.} The image-free \emph{event-prior} assigns each test tile the damaged-tile fraction estimated from training tiles of the same event. It is IID-only because a held-out event has no source labels. A tabular probe instead excludes event identity and receives hazard type, sensor, source, and ground sampling distance. \textbf{(2) Temporal view.} Probes receive $z^{\rm pre}$, $z^{\rm post}$, signed change $z^{\rm post}-z^{\rm pre}$, or complete $z_i$. Pre-only is the strict shortcut test because it contains no observed impact; post-only and signed change may contain legitimate damage evidence. \textbf{(3) Annotation density.} An image-free probe receives only $\log(1+n_i)$, where $n_i$ is annotated building count. Since $y_i=1$ when any building is damaged, larger counts create more chances for a positive tile. No building segmentation is performed. \textbf{(4) Pair correspondence.} At LOEO test time, we replace either feature with another tile's feature from the same event and label. This breaks location matching without reversing time, and reuses the probe fitted on correct source pairs. Target labels construct this stress test only, never training.
\paragraph{What shortcuts are available?}
Figure~\ref{fig:protocol} summarizes our four diagnostic families and source-only transfer evaluation. \textbf{(1) Event/context.} Can event or acquisition information predict damage without images? The \emph{event-prior} assigns each test tile its event's proportion of damaged training tiles. It is IID-only because unseen events have no training labels. A separate metadata probe receives hazard type, sensor, source, and ground sampling distance (pixel size), excluding event identity. \textbf{(2) Temporal view.} Which before/after inputs help? Separate probes receive pre-only, post-only, difference-only ($z^{\rm post}-z^{\rm pre}$), or complete $z_i$ features. Pre-only prediction observes no disaster impact; post-only and difference features may contain direct damage evidence. Differences are between embeddings, not pixels. \textbf{(3) Building count.} Can building count predict the tile label? An image-free probe receives only $\log(1+n_i)$, where $n_i$ is the annotated building count for that tile. More buildings create more opportunities for a positive label because $y_i=1$ when any building is damaged. No segmentation is performed; counts are not image-model inputs. \textbf{(4) Pair correspondence.} How much does location matching matter? At LOEO test time, we shuffle pre- or post-event features within the same event and label, then rebuild $z_i$. The probe remains trained on correct source pairs. This breaks location matching without reversing time; we average both shuffle directions. Target labels construct this diagnostic only, never training or normal prediction.

\paragraph{Source-only controls and mitigation}
% We first reweight the raw probe so that every observed event--damage label group has equal total source weight. 
We first train a classifier on the original embeddings with sample weights that give equal importance to each event--label group, preventing larger events or more frequent labels from dominating training.
We then compare four transformations fitted on source events: PCA to 256 dimensions; a two-layer MLP trained with binary cross-entropy (CE-MLP); ordinary SupCon, which clusters samples with the same damage label; and cross-event SupCon, which permits same-label positives only from different source events. Both contrastive variants use a 256-D normalized projection. Their purpose is to improve prediction on the held-out event without target-event samples. Comparing ordinary and cross-event SupCon tests whether restricting positive pairs to different source events adds a transfer benefit.

% \paragraph{Improving unseen-event prediction}
% All controls use source events only. \emph{Event--class weighting} gives each observed event--damage-label group equal total training weight. We also compare PCA compression to 256 dimensions, a supervised multilayer perceptron (MLP) trained with binary cross-entropy, and supervised contrastive learning (SupCon) \citep{Khosla2020SupCon}, which brings same-label features together. Cross-event SupCon restricts these same-label pairs to different source events. The learned features are then evaluated with a damage probe. These tests measure transfer improvements, not whether a particular shortcut has been removed.

\section{Experiments}

\paragraph{Data and implementation}
Our xView2 subset contains 9,168 optical pre/post pairs from 19 events; 35.3\% of tiles contain at least one building labeled minor damage, major damage, or destroyed. BRIGHT contains 3,029 pairs from 11 events and five hazard groups; 26.6\% contain a damaged or destroyed building. Other tiles are negative. BRIGHT pairs pre-event RGB optical imagery with post-event SAR, repeated across three channels here. This is a compatibility test, not a sensor-optimized SAR pipeline.

Images are resized to $224\times224$. TerraMind, DOFA, and DINOv3 ViT-L/16 (LVD-1689M) use model-specific RGB normalization. Prithvi-EO-2.0-300M uses display RGB range-matched to HLS RGB statistics; this does not reconstruct physical reflectance, so its results describe an out-of-domain input setting. We concatenate mean-pooled features from four backbone depths: 4,096 dimensions per image for Prithvi and DINOv3, and 3,072 for TerraMind and DOFA. Backbone weights stay fixed. Learned transformations use a 512-D hidden layer, 256-D output, AdamW, 30 epochs, and three seeds. We report balanced accuracy (BA: mean recall across the two classes) and area under the ROC curve (AUC). Learned-method results average seeds before backbones.

\paragraph{Event and context results}
Table~\ref{tab:audit}(a) shows why random-split performance is an optimistic estimate of unseen-disaster performance. Full-pair BA drops from 0.807 to 0.638 on xView2 and from 0.818 to 0.568 on BRIGHT under LOEO. Event-prior AUC reaches 0.821 and 0.715 without images. Metadata is also predictive within familiar events, but its LOEO BA is approximately chance (0.498 and 0.500). These baselines reveal available context signals; they do not establish that the image probe uses them.

\begin{table}[t]
\caption{Audit means over four backbones: (a) event/context, (b) image inputs, and (c) count/pairing controls. Full pair combines pre, post, and their difference. Shuffled pair averages both shuffle directions. Event prior is random-split only.}
\label{tab:audit}
\centering
\small
\setlength{\tabcolsep}{4pt}
\begin{tabular}{lrrrrr}
\toprule
\multicolumn{6}{l}{\textit{(a) Event and acquisition context}}\\
& \multicolumn{2}{c}{Full pair} & Event prior & \multicolumn{2}{c}{Metadata}\\
\cmidrule(lr){2-3}\cmidrule(lr){4-4}\cmidrule(lr){5-6}
Dataset & Random BA & LOEO BA & Random AUC & Random BA & LOEO BA\\
\midrule
xView2 & .807 & .638 & .821 & .725 & .498\\
BRIGHT & .818 & .568 & .715 & .600 & .500\\
\bottomrule
\end{tabular}

\smallskip
\begin{tabular}{lrrrr}
\toprule
\multicolumn{5}{l}{\textit{(b) Before/after image inputs (LOEO BA / AUC)}}\\
Dataset & Pre only & Post only & Difference only & Full pair\\
\midrule
xView2 & .577 / .611 & .617 / .665 & .587 / .626 & .638 / .699\\
BRIGHT & .571 / .612 & .553 / .587 & .561 / .606 & .568 / .606\\
\bottomrule
\end{tabular}

\smallskip
\begin{tabular}{lrrr}
\toprule
\multicolumn{4}{l}{\textit{(c) Building count and location matching (LOEO BA)}}\\
Dataset & Building count & Full pair & Shuffled pair\\
\midrule
xView2 & .629 & .638 & .617\\
BRIGHT & .653 & .568 & .552\\
\bottomrule
\end{tabular}
\end{table}

\paragraph{Image inputs and pairing}
On xView2, post-only BA exceeds pre-only (0.617 versus 0.577), and the full pair performs best at 0.638 (Table~\ref{tab:audit}b). On BRIGHT, pre-only has the highest mean BA (0.571), narrowly above the full pair (0.568). Combined inputs therefore help on xView2 but not on average in our BRIGHT pipeline, highlighting the challenge of exploiting optical--SAR pairs.

Building count alone reaches BA 0.629 and 0.653 (Table~\ref{tab:audit}c). Like pre-only success, this reveals predictive scene information, not necessarily harmful shortcut use. Shuffling pairs reduces BA from 0.638 to 0.617 on xView2 and from 0.568 to 0.552 on BRIGHT. Correct matching helps modestly when event and class are preserved.

\paragraph{Transfer improvements}
We next test source-only controls and transformations. Event--class weighting barely changes BA (Table~\ref{tab:controls}). Feature transformations give larger mean gains: SupCon performs best on xView2 (BA 0.664, AUC 0.730), and PCA on BRIGHT (0.594, 0.650). Cross-event SupCon has slightly lower mean scores than ordinary SupCon on both datasets, giving no observed benefit from restricting positive pairs to different events.

\begin{table}[t]
\caption{Source-only LOEO transfer (BA / AUC). Scores average events and backbones, after seeds for learned transformations. Bold marks the highest mean; rankings are descriptive.}
\label{tab:controls}
\centering
\small
\setlength{\tabcolsep}{8pt}
\begin{tabular}{lrr}
\toprule
Method & xView2 BA / AUC & BRIGHT BA / AUC\\
\midrule
Raw features & .638 / .699 & .568 / .606\\
Event--class weighting & .638 / .700 & .570 / .616\\
PCA-256 & .654 / .727 & \textbf{.594 / .650}\\
Supervised MLP & .658 / .714 & .582 / .615\\
SupCon & \textbf{.664 / .730} & .588 / .643\\
Cross-event SupCon & .661 / .727 & .585 / .642\\
\bottomrule
\end{tabular}
\end{table}

\section{Conclusion}
High accuracy on familiar disasters does not guarantee reliable prediction on a new disaster. Our experiments show that random splits hide shortcuts in GFM embeddings. Event statistics, scene information, and building count can predict tile-level damage, while correct pre/post matching provides a modest benefit. Feature transformations improve unseen-event performance, although cross-event SupCon does not outperform ordinary SupCon or PCA. Damage models should therefore use held-out-event evaluation and image-free, temporal-view, and correspondence-breaking controls. Future work should evaluate building-level severity and native sensor inputs.

\section{Acknolwedgement}
Computational resources were provided by the National Academic Infrastructure for Supercomputing in Sweden (NAISS), funded by the Swedish Research Council.

\bibliographystyle{plainnat}
\bibliography{main}

\end{document}